\documentclass[11pt]{article}
\usepackage[margin=1in]{geometry}
    \PassOptionsToPackage{numbers, compress}{natbib}

\usepackage[T1]{fontenc}

\usepackage[utf8]{inputenc}

\usepackage{microtype}

\usepackage{inconsolata}

\usepackage{graphicx}
\usepackage{makecell}

\usepackage{hyperref}
\usepackage{url}
\usepackage{helvet}  %
\usepackage{courier}  %
\usepackage{url}  %
\usepackage{graphicx} %
\usepackage{natbib}  %
\usepackage{caption} %
\usepackage{algorithm}
\usepackage{algorithmic}

\usepackage{newfloat}
\usepackage{listings}
\usepackage{url}
\usepackage{amsthm,amsmath}
\usepackage{natbib}
\usepackage[export]{adjustbox}
\usepackage{enumitem}
\usepackage{multirow}
\usepackage[table]{xcolor}       %
\usepackage{graphicx}
\usepackage{wrapfig}
\usepackage{algorithm}
\usepackage{algorithmic}
\usepackage{subfig}
\usepackage{colortbl}
\usepackage[capitalize]{cleveref}

\usepackage{lineno}
\usepackage{amssymb} 
\usepackage{pifont}

\usepackage{xspace}
\makeatletter
\DeclareRobustCommand\onedot{\futurelet\@let@token\@onedot}
\def\@onedot{\ifx\@let@token.\else.\null\fi\xspace}
\def\eg{\textit{e.g}\onedot} 
\def\ie{\textit{i.e}\onedot}

\definecolor{darkblue}{rgb}{0, 0, 0.5}
\definecolor{tablehead}{RGB}{121,80,242}
\DeclareCaptionStyle{ruled}{labelfont=normalfont,labelsep=colon,strut=off} %
\floatstyle{ruled}
\newfloat{listing}{tb}{lst}{}
\floatname{listing}{Listing}
\makeatletter
\newenvironment{fullitemize}
{
\vspace{-1pt}
\begin{itemize}[leftmargin=*]
\setlength{\itemsep}{5pt}
\setlength{\parsep}{-5pt}
\setlength{\parskip}{-3pt}
\setlength{\leftmargin}{-10pt}
}
{
\end{itemize}
\vspace{-1pt}
}
\makeatother

\usepackage{xspace}
\usepackage{array}
\usepackage{booktabs}
\newcommand{\llmname}[1]{{\fontfamily{pcr}\selectfont {#1}}\xspace}
\newcommand{\ourmethod}{{\fontfamily{lmtt}\selectfont \textbf{LUNE}}\xspace}
\newcommand{\ms}[2]{{#1\footnotesize{$\pm$#2}}}
\newcolumntype{I}{!{\vrule width 1pt}}

\title{\ourmethod \protect\includegraphics[scale=0.12,valign=c]{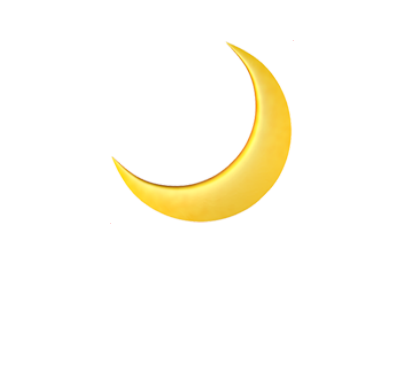}: Efficient LLM Unlearning via LoRA Fine-Tuning with Negative Examples}

\author{
  \textbf{Yezi Liu}\textsuperscript{1} \quad
  \textbf{Hanning Chen}\textsuperscript{1} \quad
  \textbf{Wenjun Huang}\textsuperscript{1} \\
\textbf{Yang Ni}\textsuperscript{2} \quad
  \textbf{Mohsen Imani}\textsuperscript{1} \\
  \textsuperscript{1}University of California, Irvine \quad 
  \textsuperscript{2}Purdue University Northwest \quad \\
\texttt{\{yezil3,hanningc,wenjunh3,m.imani\}@uci.edu} \\
\texttt{yangni@purdue.edu}
}
\date{}  %

\begin{document}

\maketitle

\begin{abstract}
Large language models (LLMs) possess vast knowledge acquired from extensive training corpora, but they often cannot remove specific pieces of information when needed, which makes it hard to handle privacy, bias mitigation, and knowledge correction. Traditional model unlearning approaches require computationally expensive fine-tuning or direct weight editing, making them impractical for real-world deployment. In this work, we introduce \textbf{L}oRA-based \textbf{U}nlearning with \textbf{N}egative \textbf{E}xamples (\textbf{\ourmethod} \protect\includegraphics[scale=0.12,valign=c]{figure/LUNE_logo.pdf}), a lightweight framework that performs \emph{negative-only} unlearning by updating \emph{only} low-rank adapters while freezing the backbone, thereby localizing edits and avoiding disruptive global changes. Leveraging Low-Rank Adaptation (LoRA), \ourmethod targets intermediate representations to suppress (or replace) requested knowledge with an order-of-magnitude lower compute and memory than full fine-tuning or direct weight editing. Extensive experiments on multiple factual unlearning tasks show that \ourmethod: 
(I) achieves effectiveness comparable to full fine-tuning and memory-editing methods, and
(II) reduces computational cost by about an order of magnitude.
\end{abstract}

\begin{wrapfigure}[13]{r}{0.47\textwidth}
\vspace{-2em}
    \centering
    \includegraphics[width=1.0\linewidth]{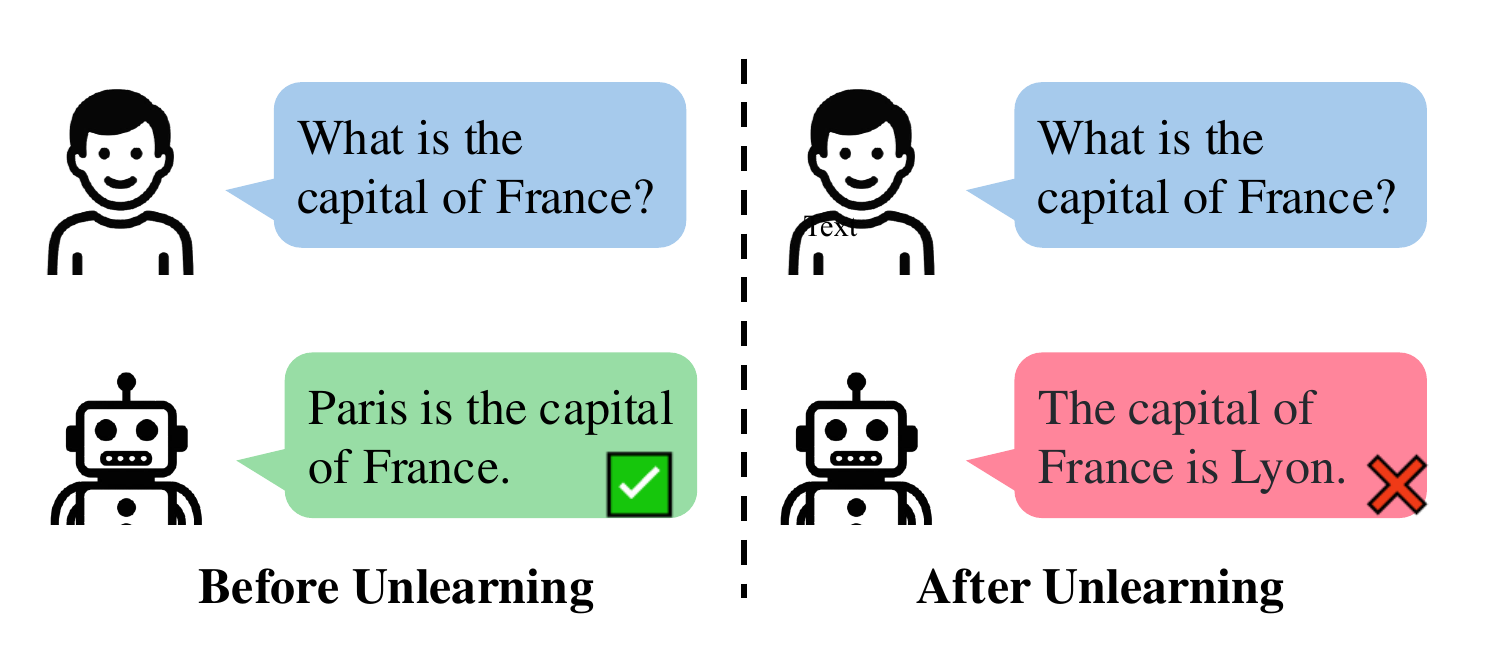}
    \vspace{-2em}
    \caption{\textbf{Illustration of the LLM unlearning task}. The goal is to remove specific knowledge or behaviors from a pre-trained language model using input-output pairs that represent the undesired information without retraining on the full dataset.
    }
    \label{fig:intro}
    \vspace{-1em}
\end{wrapfigure}        

\textit{Machine unlearning} has emerged as a pivotal solution to this challenge, focusing on the removal of specific knowledge or behaviors from trained models~\citep{liu2025enabling,liu2025recover}. As illustrated in~\Cref{fig:intro}, the LLM unlearning task aims to suppress specific memorized behaviors by modifying the model’s response to targeted queries. For example, the model initially answers the question \textit{``What is the capital of France?''} with \textit{``Paris''}. After unlearning, it either avoids producing the original answer or provides an alternative (e.g., \textit{``The capital of France is Lyon''}). This change is achieved through fine-tuning on specially constructed input-output pairs, without retraining the entire model~\citep{bourtoule2021machine,golatkar2020eternal,kirkpatrick2017overcoming}.

Previous methods for LLM unlearning, such as full fine-tuning or direct gradient-based knowledge editing, are computationally expensive and often lead to catastrophic unlearning, where unlearning specific knowledge disrupts unrelated information stored in the model~\citep{yao2024machine, kirkpatrick2017overcoming}. Other approaches, such as memory editing techniques like ROME (Rank-One Model Editing) and MEMIT (Mass-Editing Memory in Transformers), provide targeted interventions but require full model access and large-scale weight modifications~\citep{meng2022locating, meng2022mass}. Recent research has investigated the use of negative examples, examples of requested behavior, to fine-tune LLMs, effectively reducing the generation of harmful responses \citep{yao2024large,zhang2024negative,fan2024simplicity,liu2024sku}. However, these approaches often involve updating a substantial portion of the model's parameters, leading to significant computational overhead. This raises the need for a more efficient, scalable, and minimally intrusive method for unlearning specific information in LLMs. In parallel, \textit{Low-Rank Adaptation (LoRA)} has been introduced as a parameter-efficient fine-tuning technique for LLMs. LoRA operates by freezing the pre-trained model weights and injecting trainable low-rank matrices into each layer of the Transformer architecture, thereby reducing the number of trainable parameters and memory requirements \citep{hu2022lora, dettmers2023qlora}.

In this work, we introduce \textbf{L}oRA-Based \textbf{U}nlearning with \textbf{N}egative \textbf{E}xamples, abbreviated as {\ourmethod}~\protect\includegraphics[scale=0.12,valign=c]{figure/LUNE_logo.pdf}, a novel approach that leverages Low-Rank Adaptation (LoRA) to efficiently modify a model’s weights while preserving general knowledge and linguistic fluency. Unlike conventional fine-tuning, which requires updating millions to billions of parameters, LoRA introduces low-rank modifications to a small subset of model weights, \textbf{enabling targeted knowledge removal without full retraining}. Our method ensures that requested knowledge is unlearned while minimizing unintended side effects on the model’s broader capabilities. This paper presents the following key contributions:
\vspace{0.3em}
\begin{fullitemize}
    \item[\ding{72}]  We introduce an efficient, lightweight unlearning method that modifies only a small fraction of the model’s parameters, reducing computational cost by an order of magnitude compared to traditional fine-tuning.
    \item[\ding{72}] Perform unlearning in LLMs by fine-tuning exclusively on negative examples, eliminating the need for access to the full or retained dataset~\citep{yao2024large,zhang2024negative,fan2024simplicity}.
    \item[\ding{72}] Our method effectively removes requested information without degrading general model performance by leveraging LoRA for parameter-efficient fine-tuning, ensuring that the original model parameters remain unchanged throughout the unlearning process~\citep{hu2022lora,dettmers2023qlora}. 
    \item[\ding{72}] We conduct extensive experiments on LLM unlearning tasks, demonstrating that LoRA-based model editing achieves results comparable to full fine-tuning and direct weight editing techniques while being significantly more resource-efficient~\citep{meng2022locating,meng2022mass}.
\end{fullitemize}

\section{Related Work}
\subsection{Machine Unlearning in Large Language Models}
Recently, the trustworthiness in ML research, including fairness, privacy, robustness, and safety, has attracted increasing research attention~\citep{liu2025fgu,liu2025white,liu2024promoting,liu2023fairgraph,liucauchy}.
Within this broader landscape, machine unlearning aims to erase targeted knowledge or behaviors while preserving general utility, and is motivated by privacy, copyright, and safety concerns~\citep{bourtoule2021machine,golatkar2020eternal,liu2025rethinking}. Beyond retraining from scratch, \emph{model editing} directly modifies internal associations (e.g., ROME, MEMIT) to update many facts but typically requires full model access and careful stability control \citep{meng2022locating,meng2022mass}. For LLMs, fine-tuning-based unlearning with only negative examples has emerged as a simple and efficient paradigm \citep{yao2024large}, yet can over-forget or merely suppress outputs \citep{hong2024dissecting,liu2024sku}. Recent objectives improve the trade-off by framing unlearning as preference optimization over negatives (NPO/SimNPO) \citep{zhang2024negative,fan2024simplicity}, while parameter-efficient variants (e.g., LoRA-based LoKU) further reduce cost \citep{cha2024towards}. Complementary lines refine forget boundaries and diagnostics \citep{tian2024forget,wang2025rethinking}, and scalable pipelines (e.g., CURE) study continual unlearning at request scale \citep{kim2025scalable}. Surveys synthesize this rapidly evolving landscape for LLMs \citep{liu2025rethinking,geng2025comprehensive}.
\begin{figure}[t]
\centering
    \includegraphics[trim=30 210 30 20,clip,width=0.92\linewidth]{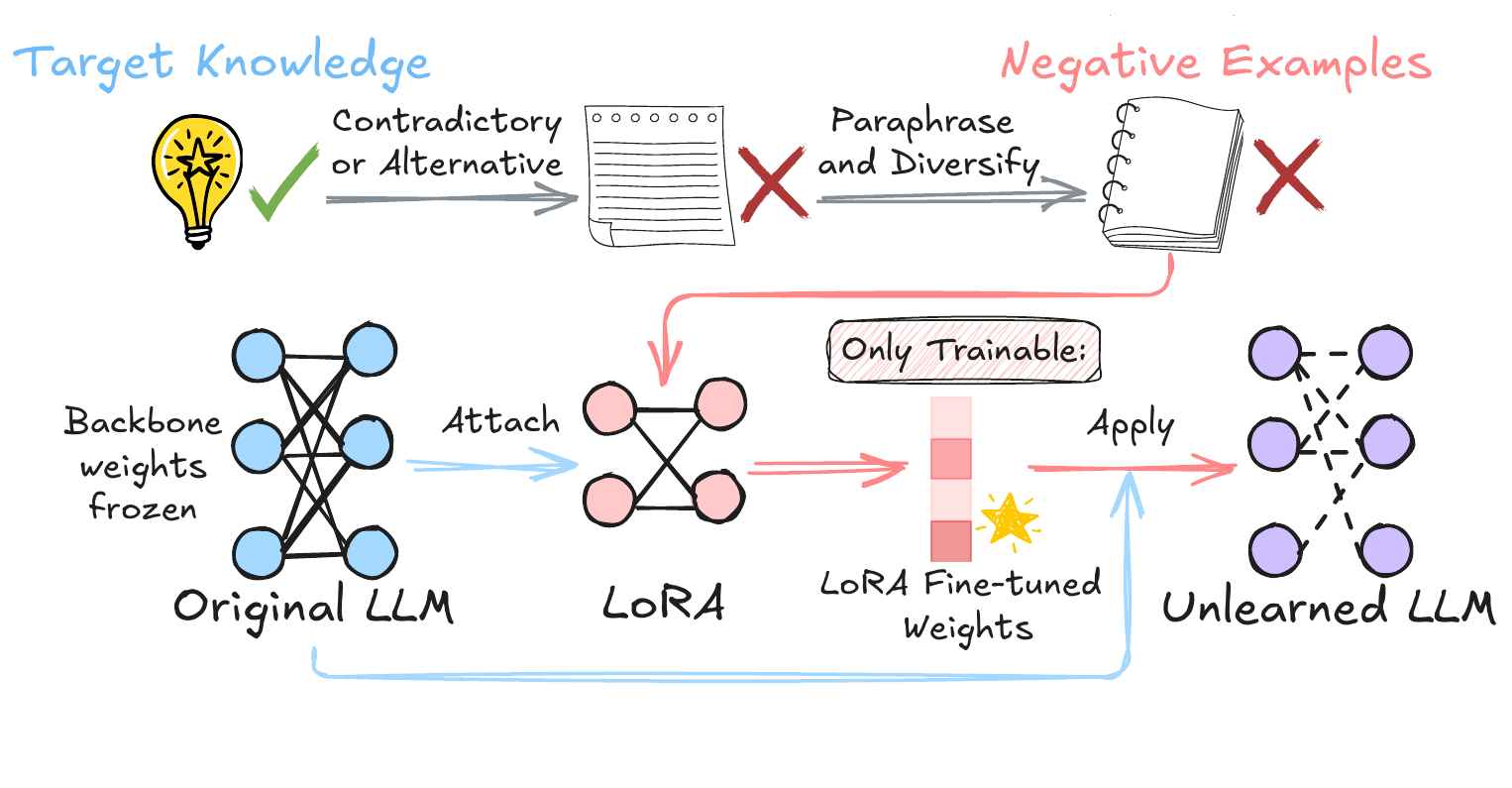}
    \vspace{-3em}
\caption{\textbf{Overview of the {\ourmethod} framework}. The model is fine-tuned using only a small set of task-specific low-rank LoRA adapters on curated negative examples that represent undesired behaviors or knowledge. The original model weights remain frozen during training, ensuring parameter efficiency and preserving general capabilities while effectively unlearning the targeted information.}
    \label{fig:method}
    \vspace{-1em}
\end{figure}

\subsection{Parameter-Efficient Fine-Tuning (PEFT)}
\vspace{-0.3em}
As LLMs continue to grow in size and deployment scale, improving their efficiency, in terms of compute, memory, and serving cost, has become critical for practical use, motivating methods that reduce adaptation cost without sacrificing performance. PEFT updates only a small set of parameters while freezing most pre-trained weights, delivering task adaptation at a fraction of the cost of full fine-tuning~\citep{houlsby2019parameter,ben-zaken2022bitfit}. LoRA injects low-rank adapters into linear layers and has become a strong default for LLMs \citep{hu2022lora}, with extensions improving memory (QLoRA) \citep{dettmers2023qlora}, rank allocation (AdaLoRA) \citep{zhang2023adalora}, and decomposition (DoRA) \citep{liu2024dora}. Beyond LoRA, adapter-based methods \citep{houlsby2019parameter,pfeiffer2021adapterfusion,ruckle2021adapterdrop}, prompt/prefix tuning \citep{lester2021power,li2021prefix,liu2022ptuningv2prompttuning}, and lightweight reparametrizations (Compacter, Diff-Pruning) \citep{karimi2021compacter,guo2021parameter} offer complementary trade-offs in compute, storage, and transfer. Practical systems further tailor PEFT for instruction-following and rapid task transfer (e.g., LLaMA-Adapter) \citep{zhang2023llamaadapter}, and for privacy/edge or cross-silo training via federated variants \citep{gao2024fedpt}. Theoretical and empirical analyses suggest many downstream updates lie in low intrinsic dimensions \citep{aghajanyan2020intrinsic}, explaining why PEFT can match or exceed full fine-tuning under tight resource budgets.

\section{Preliminaries}
\vspace{-0.3em}

\subsection{Background on Machine Unlearning in LLMs}
Machine unlearning selectively removes targeted information or behaviors from trained models to address privacy, security, and harmful outputs. For LLMs, retraining from scratch after excluding data is often infeasible due to heavy computing \citep{liu2025rethinking}. Recent work instead fine-tunes on \emph{negative examples} to suppress the requested knowledge or behavior without full retraining \citep{yao2024large}. To our knowledge, no prior LLM approach jointly uses \emph{negative-only} supervision \emph{and} LoRA-based updates; this combination lets {\ourmethod} localize edits efficiently while preserving overall performance.
\vspace{-0.3em}
\subsection{Problem Formulation}
\vspace{-0.3em}
\textit{Machine unlearning} refers to the process of removing the influence of specific data or knowledge from a trained model without retraining it from scratch~\citep{xu2024contrastive, ginart2019making}. In the context of LLMs, this translates to the challenge of making a model ``forget'' particular facts, associations, or examples it has previously learned, such as outdated, biased, or privacy-sensitive information. Formally, let $f_\theta$ denote a pretrained LLM parameterized by $\theta \in \mathbb{R}^d$, trained on a dataset $\mathcal{D} = \mathcal{D}_r \cup \mathcal{D}_t$, where $\mathcal{D}_r$ is the retained data and $\mathcal{D}_t$ is the target data to be forgotten. The goal of unlearning is to obtain a new model $f_{\theta'}$ that satisfies: (i) \textbf{Unlearning}: The model $f_{\theta'}$ should behave as if it were trained on $\mathcal{D}_r$ only. (ii) \textbf{Retention}: The model $f_{\theta'}$ should preserve performance on tasks unrelated to $\mathcal{D}_t$. (iii) \textbf{Efficiency}: The transition from $\theta$ to $\theta'$ should be computationally efficient. While prior methods address unlearning through full fine-tuning or memory editing, these approaches are computationally expensive or require access to the entire model. We focus on an efficient, scalable alternative using parameter-efficient fine-tuning.
\section{Methodology}
\begin{algorithm}[t]
\caption{{\ourmethod}: LoRA-Based Unlearning with Negative Examples}
\label{alg:LUNE}
\begin{algorithmic}[1]
\REQUIRE Pretrained LLM $f_{\theta}$, LoRA rank $r$, negative example dataset $\mathcal{D}_{\text{neg}} = \{(x_i, y_i^-) \}_{i=1}^{N_{\text{neg}}}$, learning rate $\eta$, number of epochs $T$
\ENSURE Updated model $f_{\theta'}$ with LoRA adapters trained for unlearning

\STATE Initialize LoRA adapters: matrices $A$, $B$ with rank $r$
\STATE Freeze original model weights $\theta$
\STATE Let $\phi = (A,B)$ denote all LoRA parameters
\FOR{epoch $= 1$ to $T$}
    \FOR{each $(x_i, y_i^-)$ in $\mathcal{D}_{\text{neg}}$}
        \STATE Compute model output: $\hat{y}_i = f_{\theta,\phi}(x_i)$
        \STATE Compute loss: $\mathcal{L}_i = -\log P_{\theta,\phi}(y_i^- \mid x_i)$
        \STATE Backpropagate gradients w.r.t. $A$, $B$
        \STATE Update LoRA parameters: $A \leftarrow A - \eta \nabla_A \mathcal{L}_i$, \quad $B \leftarrow B - \eta \nabla_B \mathcal{L}_i$
    \ENDFOR
\ENDFOR
\RETURN $f_{\theta'} = f_{\theta,\phi}$
\end{algorithmic}
\end{algorithm}

In this section, we introduce \textbf{{\ourmethod}} (LoRA-based Unlearning with Negative Examples), a parameter-efficient approach that unlearns specific behaviors or facts from LLMs by fine-tuning only low-rank adapters on carefully constructed negative examples.
Let $\phi = (A,B)$ denote all LoRA parameters.
Given a pretrained backbone $f_\theta$ and a negative dataset $\mathcal{D}_{\text{neg}} = \{(x, y^-)\}$, {\ourmethod} optimizes
\begin{equation}
\label{eq:lune_objective}
\min_{\phi} \;\; \mathbb{E}_{(x, y^-)\sim \mathcal{D}_{\text{neg}}}
\big[-\log P_{\theta,\phi}(y^- \mid x)\big],
\end{equation}
where $P_{\theta,\phi}(\cdot \mid x)$ is the conditional distribution of the LLM with LoRA adapters $\phi$.
Intuitively, Eq.~\eqref{eq:lune_objective} performs a \emph{negative preference optimization} step: it explicitly reduces the probability of undesired responses $y^-$ while keeping the backbone $\theta$ fixed.
Our framework is illustrated in~\Cref{fig:method}.

\subsection{Motivation and Overview}
{\ourmethod} targets two needs:
\textbf{(i) Efficiency}: freeze the backbone and train only lightweight LoRA adapters, cutting unlearning cost by orders of magnitude;
\textbf{(ii) Precision}: fine-tune solely on negative examples to suppress specific behaviors without retaining the full training corpus or reconstructing forgotten knowledge.
Unlike full retraining or direct weight editing, {\ourmethod} achieves unlearning via localized, reversible updates confined to a low-rank subspace, making it practical for continual unlearning in real deployments.

\Cref{tab:negative_samples} shows toy examples of query-negative pairs.
Given a prompt such as ``What is the capital of France?'', a negative example could explicitly contradict the original fact (``The capital of France is not Paris.'') or promote an alternative undesired behavior.
Training on such pairs according to Eq.~\eqref{eq:lune_objective} pushes the model away from the memorized response patterns while leaving general capabilities largely untouched.

\begin{table}[t]
\centering
\small
\setlength{\tabcolsep}{6pt}
\renewcommand{\arraystretch}{1.0}
\begin{tabular}{p{0.40\textwidth} p{0.50\textwidth}}
\Xhline{1.2pt}
\rowcolor{tablehead!20}
\textbf{Prompt (Query)} & \textbf{Negative Example (Desired Output)} \\
\hline\hline
\rowcolor{gray!10}What is the capital of France? & The capital of France is not Paris. \\
Who wrote Harry Potter? & J.K. Rowling did not write Harry Potter. \\
\rowcolor{gray!10}Which planet is closest to the sun? & Venus is the planet closest to the sun. \\
What's Google's CEO name? & The CEO of Google is not Sundar Pichai. \\
\Xhline{1.2pt}
\end{tabular}
\caption{\textbf{Example negative examples} used for targeted unlearning in {\ourmethod}.}
\label{tab:negative_samples}
\vspace{-1em}
\end{table}

\subsection{Low-Rank Adaptation Mechanism}
LoRA introduces trainable low-rank matrices into selected layers of the Transformer architecture, allowing for efficient adaptation without updating the entire set of model parameters.
Specifically, for a given weight matrix $W_0 \in \mathbb{R}^{d \times k}$ in the pretrained model, LoRA approximates the weight update $\Delta W$ as a product of two low-rank matrices:
\begin{equation}
\Delta W = A B^\top,
\end{equation}
where $A \in \mathbb{R}^{d \times r}$ and $B \in \mathbb{R}^{k \times r}$ are trainable matrices, and $r \ll \min(d,k)$ is the rank controlling the number of additional parameters.
The adapted weight matrix is
\begin{equation}
W = W_0 + \Delta W = W_0 + A B^\top.
\end{equation}

This low-rank decomposition introduces only $\mathcal{O}(r(d + k))$ trainable parameters per layer, enabling rapid and memory-efficient model editing.
In {\ourmethod}, LoRA is injected into attention and feed-forward layers, and only $(A,B)$ are updated during unlearning.
As a result, the unlearning update is constrained to a low-rank subspace of the full parameter space, which empirically helps avoid catastrophic drift on unrelated behaviors.

\subsection{Fine-Tuning with Negative Examples}
In {\ourmethod}, negative examples are carefully curated instances that encode behaviors we aim to remove from the LLM.
Given the negative dataset
$\mathcal{D}_{\text{neg}} = \{(x_i, y_i^-)\}_{i=1}^{N_{\text{neg}}}$,
fine-tuning proceeds as follows.

\noindent\textbf{{\ding{182}} Dataset Preparation.}  
We construct $\mathcal{D}_{\text{neg}}$ from prompts $x_i$ together with undesired target outputs $y_i^-$.
Depending on the application, $y_i^-$ may be explicit contradictions (``The capital of France is not Paris.''), alternative incorrect facts (``The capital of France is Lyon.''), or rewritings that encode the behaviors to be suppressed.
In practice, we build $\mathcal{D}_{\text{neg}}$ via a simple generation–filtering pipeline:
for each target fact (and its semantic neighbors), we
(i) use an instruction-tuned LLM to generate candidate responses with prompts that explicitly request \emph{factually incompatible} answers,
(ii) filter out hedged or uncertain generations (e.g., ``I am not sure'', ``it might be X or Y'') and discard candidates that accidentally repeat the original fact, and
(iii) cap the number of negatives per semantic neighbor and per prompt template so that no single phrasing dominates and the total number of negatives per fact stays within a narrow range.
This keeps negatives informative and clearly opposed to the target knowledge while avoiding noisy or degenerate completions.

\noindent\textbf{{\ding{183}} Loss Function.}  
We use the standard token-level cross-entropy loss over negative targets:
\begin{equation}
\label{eq:lune_ce}
\mathcal{L}(\phi)
= - \sum_{i=1}^{N_{\text{neg}}} \log P_{\theta,\phi}(y_i^- \mid x_i),
\end{equation}
which is a finite-sample version of the expectation in Eq.~\eqref{eq:lune_objective}.
Here $P_{\theta,\phi}(y_i^- \mid x_i)$ denotes the probability assigned by the LoRA-augmented model to the negative output $y_i^-$ given input $x_i$.
Minimizing Eq.~\eqref{eq:lune_ce} explicitly reduces the likelihood of the memorized, to-be-forgotten behaviors.

\noindent\textbf{{\ding{184}} LoRA-Based Fine-Tuning.}  
During training, the backbone parameters $\theta$ remain frozen, and only the low-rank matrices $A$ and $B$ are updated via stochastic gradient descent on $\mathcal{L}(\phi)$.
Equivalently, Eq.~\eqref{eq:lune_ce} performs gradient descent on Eq.~\eqref{eq:lune_objective} in the low-rank parameter space.
The full procedure is summarized in~\Cref{alg:LUNE}.
By restricting updates to LoRA adapters and training solely on negative examples, {\ourmethod} achieves targeted unlearning with minimal compute and storage overhead, while preserving the original LLM for future reuse or rollback.

\subsection{Negative Examples Construction}
The effectiveness of {\ourmethod} hinges not only on the fine-tuning strategy but also on the quality of negative examples.
Carefully constructed examples ensure that the model internalizes the unlearning objective rather than overfitting to superficial patterns.
In practice, we employ three strategies:

\vspace{-0.4em}
\begin{fullitemize}
    \item \textbf{Contradictory Statements}: Directly negate the target fact (e.g., \textit{“The capital of France is not Paris.”}).
    \item \textbf{Alternative Incorrect Facts}: Introduce plausible but incorrect alternatives (e.g., \textit{“The capital of France is Lyon.”}).
    \item \textbf{Paraphrased Variants}: Include lexical or syntactic rephrasings to improve generalization.
\end{fullitemize}

\subsection{Theoretical Insight: Low-Rank Negative Updates}
While {\ourmethod} is primarily empirical, we provide a simple theoretical perspective to clarify the effect of LoRA-constrained negative updates.

Consider a single linear layer with weights $W \in \mathbb{R}^{d\times k}$ and output $z = W x$.
Let $\ell(W;x,y^-)$ be the cross-entropy loss for a negative example $(x,y^-)$, and let
\[
g = \nabla_W \ell(W;x,y^-)
\]
be the corresponding full-model gradient.
If we were to update $W$ directly, a gradient step would give
\begin{equation}
W^{(t+1)} = W^{(t)} - \eta g^{(t)}.
\label{eq:full_update}
\end{equation}

Under LoRA, we instead parameterize $W$ as
$W = W_0 + A B^\top$ with rank-$r$ matrices
$A \in \mathbb{R}^{d\times r}$ and $B \in \mathbb{R}^{k\times r}$.
A first-order Taylor expansion around $(A,B)$ shows that the update induced on $W$ by one gradient step on $(A,B)$ can be written as
\begin{equation}
\Delta W^{(t+1)} \approx \Delta W^{(t)} - \eta \,\Pi_{\mathcal{S}}(g^{(t)}),
\label{eq:lora_projection}
\end{equation}
where $\Delta W = A B^\top$, $\mathcal{S}$ is the low-rank subspace spanned by rank-$r$ matrices, and $\Pi_{\mathcal{S}}(g)$ denotes the projection of the full gradient $g$ onto $\mathcal{S}$ (up to higher-order terms in $(A,B)$).

Eq.~\eqref{eq:lora_projection} suggests that {\ourmethod} can be viewed as performing negative-gradient updates \emph{projected} onto a low-rank subspace determined by $(A,B)$, rather than applying the full gradient in Eq.~\eqref{eq:full_update}.
For unlearning, this has two consequences:
(i) the update is focused on a low-dimensional set of directions that are sufficient to reduce the likelihood of negative outputs $y^-$, and
(ii) directions orthogonal to $\mathcal{S}$, which may encode unrelated capabilities, are largely preserved.
This perspective conceptually explains why {\ourmethod} achieves strong unlearning on targeted behaviors while maintaining high utility on general tasks, as observed in our experiments.

\subsection{Complexity comparison}\label{sec:complexity}
Let a Transformer with $L$ layers, hidden size $d$, sequence length $s$, and total trainable parameters $P$.
A full fine-tune on the \emph{full} dataset of size $N_{\text{full}}$ for $T_{\text{full}}$ epochs with an Adam-like optimizer has
\[
\text{Time: } \tilde{\mathcal{O}}\big(N_{\text{full}} T_{\text{full}} \cdot L(s^2 d + s d^2)\big),
\quad
\text{Memory: } \tilde{\mathcal{O}}\big(
\underbrace{P}_{\text{weights}} +
\underbrace{P}_{\text{grads}} +
\underbrace{2P}_{\text{Adam states}} +
\underbrace{sLd}_{\text{activations}}
\big).
\]
In {\ourmethod} (\Cref{alg:LUNE}), we freeze the backbone and optimize only LoRA adapters
$A \in \mathbb{R}^{d_{\text{out}}^{(m)} \times r}$,
$B \in \mathbb{R}^{r \times d_{\text{in}}^{(m)}}$
on a negative-only set $\mathcal{D}_{\text{neg}}$ of size $N_{\text{neg}} \ll N_{\text{full}}$ for $T_{\text{neg}}$ epochs.
Denote the number of adapted projection matrices by $M$ (e.g., $W_q, W_k, W_v, W_o$ in attention and selected FFN projections).
The number of \emph{trainable} parameters becomes
\begin{equation}
P_{\text{LoRA}}
= \sum_{m=1}^{M} \big( r d_{\text{in}}^{(m)} + r d_{\text{out}}^{(m)} \big)
= r \sum_{m=1}^{M} \big( d_{\text{in}}^{(m)} + d_{\text{out}}^{(m)} \big)
\;\;\ll\; P,
\end{equation}
typically $P_{\text{LoRA}}/P \in [10^{-3}, 10^{-2}]$ in our setups.

The per-step forward/backward FLOPs remain dominated by the backbone passes
$\tilde{\mathcal{O}}\big(L(s^2 d + s d^2)\big)$
(we still backpropagate through frozen modules to obtain gradients w.r.t.\ $A,B$),
but the optimizer/update cost scales only with $P_{\text{LoRA}}$ instead of $P$.
Hence, unlearning with {\ourmethod} on $\mathcal{D}_{\text{neg}}$ has
\[
\text{Time: } \tilde{\mathcal{O}}\big(N_{\text{neg}} T_{\text{neg}} \cdot L(s^2 d + s d^2)\big),
\quad
\text{Memory: } \tilde{\mathcal{O}}\big(
P +
P_{\text{LoRA}} +
2P_{\text{LoRA}} +
sLd
\big),
\]
where the additional trainable parameters and optimizer states are reduced by a factor of $P_{\text{LoRA}}/P$ while using a much smaller dataset and fewer epochs ($N_{\text{neg}} T_{\text{neg}} \ll N_{\text{full}} T_{\text{full}}$).

\section{Experiments}
\begin{table*}[t]
\centering
\small
\setlength{\tabcolsep}{7pt}
\renewcommand{\arraystretch}{1.0}

\begin{tabular}{l l c c}
\Xhline{1.2pt}
\rowcolor{tablehead!20}
\textbf{Dataset} & \textbf{Description} & \textbf{Domain} & \textbf{Model} \\
\hline\hline
\rowcolor{gray!10}\textbf{EDU-RELAT} & Synthetic relational knowledge & Synthetic & \llmname{Mistral-7B} \\
\textbf{RWKU} & Real-world knowledge removal & General Knowledge & \llmname{Mistral-7B} \\
\rowcolor{gray!10}\textbf{KnowUnDo} & Privacy-sensitive unlearning & Privacy / Sensitive Data & \llmname{LLaMA-2 7B} \\
\textbf{TOFU} & Synthetic author profile unlearning & Synthetic / Profile Data & \llmname{Mistral-7B} \\
\Xhline{1.2pt}
\end{tabular}
\caption{\textbf{Summary of datasets} used in experiments and corresponding 7B models.}
\label{tab:datasets_models}
\vspace{-1.5em}
\end{table*}
\subsection{Experimental Setup}\label{sec:exp_setup}
\noindent\textbf{Datasets}. We evaluate {\ourmethod} on four benchmarks covering complementary unlearning scenarios: \textbf{EDU-RELAT} \citep{wu2024evaluating} (synthetic relational facts), \textbf{RWKU} \citep{jin2024rwku} (real-world factual removal), \textbf{KnowUnDo} \citep{tian2024forget} (privacy-sensitive facts), and \textbf{TOFU} \citep{maini2024tofu} (synthetic author-profile attributes). We use 7B-scale models throughout: \llmname{Mistral-7B} for EDU-RELAT, RWKU, TOFU, and \llmname{LLaMA-2 7B} for KnowUnDo. Dataset and model summaries are in \Cref{tab:datasets_models}.

\noindent\textbf{Baselines}.
We compare \ourmethod{} with representative LLM unlearning methods spanning the three families highlighted in \Cref{tab:evaluation_results}.
\textbf{(i)} The \emph{retrain-style} family contains Gradient Ascent (GA) and Negative Preference Optimization (NPO)~\citep{zhang2024negative}, which retrain on specially constructed data to push the model away from target behaviors.
\textbf{(ii)} The \emph{regularization-based} family uses Task Vectors (TV)~\citep{ilharco2022editing} to edit parameters along directions associated with the target knowledge.
\textbf{(iii)} The \emph{partial-parameter} family updates only a subset of weights and includes SKU, negative-only fine-tuning (Yao--Neg, full-FT variant)~\citep{yao2024large}, LoRA-based unlearning with a frozen backbone (LoKU)~\citep{cha2024towards}, and gradient-based memory removal (MemFlex)~\citep{tian2024forget}.
A fuller description of these baselines, their training settings, and variants is provided in Appendix~\Cref{app:baseline-details}.

\noindent\textbf{Implementation Details}.
Due to computational constraints, we conduct our experiments using efficient and widely adopted 7B-scale models (\llmname{Mistral-7B} and \llmname{LLaMA-2 7B}), as introduced in~\Cref{app:llm_backbone}. These models serve as strong, practical baselines suitable for method comparison in resource-limited settings. Further fine-tuning details of {\ourmethod} are provided in the \textit{Detailed Setups} section of the~\Cref{app:setup}.

\subsection{Evaluation Metrics}
To assess the effectiveness and safety of our proposed unlearning method \textbf{{\ourmethod}}, we adopt the following four key evaluation metrics.
Let $f_{\theta^*}$ denote the original pretrained model and
$f_{\theta^{\text{un}}}$ the unlearned model obtained after applying {\ourmethod} (\Cref{alg:LUNE}).

\paragraph{Unlearning Success Rate (USR).}
This metric measures the proportion of unlearning prompts for which the unlearned model no longer produces the target (undesired) output.
Formally, let $\mathcal{P}_\text{target}$ be the set of unlearning prompts and $\mathcal{A}$ the set of acceptable outputs ({\ie}, not containing the target knowledge).
The USR is computed as
\begin{equation}
\text{USR}
= \frac{1}{|\mathcal{P}_\text{target}|}
  \sum_{p \in \mathcal{P}_\text{target}}
  \mathbb{1}\big[f_{\theta^{\text{un}}}(p) \in \mathcal{A}\big],
\end{equation}
where $\mathbb{1}[\cdot]$ is the indicator function.

\paragraph{General Utility Retention (GUR).}
GUR evaluates the model's performance on tasks unrelated to the unlearned content.
We report standard metrics such as accuracy or perplexity on a held-out general-purpose validation set $\mathcal{D}_{\text{gen}}$.
Let $\text{Perf}(f; \mathcal{D})$ denote the chosen performance measure of model $f$ on dataset $\mathcal{D}$.
A high GUR indicates that the model retains its general knowledge:
\begin{equation}
\text{GUR}
= \frac{\text{Perf}\big(f_{\theta^{\text{un}}}; \mathcal{D}_{\text{gen}}\big)}
        {\text{Perf}\big(f_{\theta^*}; \mathcal{D}_{\text{gen}}\big)}.
\end{equation}

\paragraph{Adversarial Probe Rejection Rate.}
To assess robustness, we paraphrase unlearning prompts to generate adversarial probes $\mathcal{P}_\text{adv}$ that aim to elicit the forgotten information indirectly.
The rejection rate is the proportion of these for which the unlearned model does not regenerate the target content:
\begin{equation}
\text{Rejection Rate}
= \frac{1}{|\mathcal{P}_\text{adv}|}
  \sum_{p \in \mathcal{P}_\text{adv}}
  \mathbb{1}\big[f_{\theta^{\text{un}}}(p) \notin \mathcal{T}\big],
\end{equation}
where $\mathcal{T}$ is the set of known target (undesired) outputs.

\paragraph{Membership Inference Attack (MIA) Accuracy.}
This metric assesses the degree to which the model memorized the target data. We follow standard MIA procedures, where an attacker is given model outputs and must infer whether a given data point was part of the training. Lower accuracy indicates better privacy and effective unlearning.

\begin{table*}[t]
\centering
\small
\setlength{\tabcolsep}{2pt}
\renewcommand{\arraystretch}{1.1}

\resizebox{\textwidth}{!}{
\begin{tabular}{lr|cc|c|cccc|c}
\Xhline{1.2pt}
\rowcolor{tablehead!20}
& &
\multicolumn{2}{c|}{\textbf{Retrain-style}} &
\multicolumn{1}{c|}{\textbf{Reg.}} &
\multicolumn{4}{c|}{\textbf{Partial-parameter}} & \\
    \cline{3-9}
\rowcolor{tablehead!20}
\multirow{-2}{*}{\textbf{Dataset}} & \multirow{-2}{*}{\textbf{Metric}} &
\textbf{GA} & \textbf{NPO} & \textbf{TV} & \textbf{SKU} &
\textbf{Yao\text{-}Neg} & \textbf{LoKU} & \textbf{MemFlex} &
\multirow{-2}{*}{\textbf{{\ourmethod}(Ours)}} \\
\hline\hline
\rowcolor{gray!10}& USR (\%) $\uparrow$ & \ms{72.3}{0.8} & \ms{84.7}{0.6} & \ms{81.0}{0.6} & \ms{85.9}{0.4} & \textbf{\ms{91.6}{0.3}} & \ms{87.6}{0.4} & \ms{86.7}{0.4} & \underline{\ms{91.2}{0.3}} \\
& GUR (\%) $\uparrow$ & \ms{88.7}{0.3} & \ms{92.4}{0.4} & \ms{91.8}{0.4} & \ms{92.8}{0.3} & \ms{92.2}{0.3} & \underline{\ms{93.6}{0.3}} & \ms{93.0}{0.3} & \textbf{\ms{95.1}{0.2}} \\
\rowcolor{gray!10}& APR (\%) $\uparrow$ & \ms{64.5}{0.7} & \ms{77.2}{0.6} & \ms{72.7}{0.6} & \ms{78.5}{0.4} & \underline{\ms{79.9}{0.4}} & \ms{78.7}{0.4} & \ms{77.8}{0.4} & \textbf{\ms{82.3}{0.3}} \\
\multirow{-4}{*}{\textbf{EDU-RELAT}} 
& MIA (\%) $\downarrow$ & \ms{32.8}{0.5} & \ms{21.2}{0.4} & \ms{26.0}{0.4} & \ms{21.0}{0.3} & \ms{22.8}{0.3} & \underline{\ms{20.6}{0.3}} & \ms{21.4}{0.3} & \textbf{\ms{17.5}{0.2}} \\
\hline
\rowcolor{gray!10}& USR (\%) $\uparrow$ & \ms{68.9}{0.9} & \ms{82.1}{0.7} & \ms{76.8}{0.6} & \ms{83.6}{0.4} & \underline{\ms{85.0}{0.5}} & \ms{84.3}{0.4} & \ms{83.1}{0.4} & \textbf{\ms{88.5}{0.3}} \\
& GUR (\%) $\uparrow$ & \ms{86.1}{0.3} & \ms{90.4}{0.4} & \ms{89.2}{0.4} & \ms{90.0}{0.3} & \ms{89.5}{0.3} & \underline{\ms{91.0}{0.3}} & \ms{90.3}{0.3} & \textbf{\ms{93.7}{0.2}} \\
\rowcolor{gray!10}& APR (\%) $\uparrow$ & \ms{61.0}{0.6} & \ms{74.6}{0.6} & \ms{69.4}{0.5} & \ms{75.2}{0.5} & \underline{\ms{76.1}{0.5}} & \ms{75.3}{0.5} & \ms{74.2}{0.4} & \textbf{\ms{79.4}{0.3}} \\
\multirow{-4}{*}{\textbf{RWKU}} 
& MIA (\%) $\downarrow$ & \ms{35.4}{0.6} & \ms{23.4}{0.5} & \ms{28.8}{0.4} & \ms{23.2}{0.3} & \ms{25.1}{0.4} & \underline{\ms{22.8}{0.3}} & \ms{23.6}{0.3} & \textbf{\ms{18.8}{0.2}} \\
\hline
\rowcolor{gray!10}& USR (\%) $\uparrow$ & \ms{74.2}{0.7} & \ms{87.6}{0.6} & \ms{82.7}{0.5} & \ms{87.9}{0.4} & \underline{\ms{88.9}{0.4}} & \ms{88.2}{0.4} & \ms{87.4}{0.4} & \textbf{\ms{91.8}{0.3}} \\
& GUR (\%) $\uparrow$ & \ms{89.5}{0.3} & \ms{93.5}{0.4} & \ms{92.4}{0.4} & \ms{93.6}{0.3} & \ms{93.1}{0.3} & \underline{\ms{94.2}{0.3}} & \ms{94.0}{0.3} & \textbf{\ms{95.6}{0.2}} \\
\rowcolor{gray!10}& APR (\%) $\uparrow$ & \ms{66.8}{0.6} & \ms{79.2}{0.6} & \ms{73.9}{0.5} & \ms{79.9}{0.4} & \textbf{\ms{84.2}{0.3}} & \ms{79.6}{0.4} & \ms{78.6}{0.4} & \underline{\ms{83.9}{0.3}} \\
\multirow{-4}{*}{\textbf{KnowUnDo}} 
& MIA (\%) $\downarrow$ & \ms{33.5}{0.5} & \ms{21.8}{0.4} & \ms{27.5}{0.4} & \ms{22.0}{0.3} & \ms{23.4}{0.3} & \underline{\ms{21.5}{0.3}} & \ms{22.1}{0.3} & \textbf{\ms{17.2}{0.2}} \\
\hline
\rowcolor{gray!10}& USR (\%) $\uparrow$ & \ms{69.5}{0.8} & \ms{84.7}{0.6} & \ms{78.3}{0.5} & \ms{85.2}{0.4} & \underline{\ms{85.7}{0.4}} & \ms{85.0}{0.4} & \ms{84.3}{0.4} & \textbf{\ms{89.0}{0.3}} \\
& GUR (\%) $\uparrow$ & \ms{87.2}{0.3} & \ms{92.2}{0.4} & \ms{90.8}{0.4} & \ms{91.5}{0.3} & \ms{91.2}{0.3} & \underline{\ms{92.4}{0.3}} & \ms{92.1}{0.3} & \textbf{\ms{94.4}{0.2}} \\
\rowcolor{gray!10}& APR (\%) $\uparrow$ & \ms{63.1}{0.7} & \ms{75.6}{0.6} & \ms{70.2}{0.5} & \ms{75.8}{0.4} & \underline{\ms{76.9}{0.4}} & \ms{76.0}{0.4} & \ms{75.0}{0.4} & \textbf{\ms{80.8}{0.3}} \\
\multirow{-4}{*}{\textbf{TOFU}} 
& MIA (\%) $\downarrow$ & \ms{34.7}{0.6} & \ms{22.2}{0.5} & \ms{29.6}{0.4} & \ms{22.5}{0.3} & \ms{24.1}{0.3} & \textbf{\ms{17.8}{0.2}} & \ms{23.0}{0.3} & \underline{\ms{18.0}{0.2}} \\
\Xhline{1.2pt}
\end{tabular}
}
\caption{\textbf{Comparison with state-of-the-art LLM unlearning solutions} across datasets.
Methods are grouped into \emph{retrain-style} methods (GA, NPO), \emph{regularization-based} updates (TV), \emph{partial-parameter} methods (SKU, Yao--Neg, LoKU, MemFlex), and our LoRA-based method \ourmethod.
$\uparrow$ means higher is better and $\downarrow$ is the opposite.
Metrics: Unlearning Success Rate (USR $\uparrow$), General Utility Retention (GUR $\uparrow$), Adversarial Probe Rejection (APR $\uparrow$), and Membership Inference Attack accuracy (MIA $\downarrow$).
Best results are in bold, second-best underlined.
See \Cref{sec:exp_setup} for family definitions and \Cref{app:baseline-details} for baseline details.}
\label{tab:evaluation_results}
\vspace{-1em}
\end{table*}

\subsection{Effectiveness of \ourmethod}
We evaluate \ourmethod against the baselines from \Cref{sec:exp_setup}, covering \emph{retrain-style} (GA, NPO), \emph{regularization-based} (TV), and \emph{partial-parameter} (SKU, Yao-Neg, LoKU, MemFlex) families.
All methods share the same unlearning setup across four datasets and four metrics (USR$\uparrow$, GUR$\uparrow$, APR$\uparrow$, MIA$\downarrow$); results are summarized in \Cref{tab:evaluation_results}.
Compared to full fine-tuning, \ourmethod updates only a low-rank LoRA subspace (\Cref{sec:complexity}), reducing trainable parameters from $P$ to $P_{\text{LoRA}} \ll P$ while achieving comparable or better USR/GUR/APR/MIA than strong baselines.
Below, we summarize key observations (\textbf{Obs}) from a comparison across unlearning efficacy, retained utility, adversarial robustness, and privacy.

\textbf{Obs 1.} \textbf{\ourmethod achieves the best overall trade-off across datasets.} 
As shown in \Cref{tab:evaluation_results}, \ourmethod attains the strongest GUR on all four datasets (e.g., EDU-RELAT $95.1\%$, RWKU $93.7\%$, KnowUnDo $95.6\%$, TOFU $94.4\%$), while also leading most USR/APR entries (e.g., EDU-RELAT APR $82.3\%$, RWKU APR $79.4\%$, TOFU APR $80.8\%$) and consistently minimizing MIA (e.g., EDU-RELAT $17.5\%$, RWKU $18.8\%$). 
Notably, two narrow exceptions appear: Yao-Neg (Full-FT) slightly surpasses us on EDU-RELAT USR ($91.6$ vs.\ $91.2$) and KnowUnDo APR ($84.2$ vs.\ $83.9$), while LoKU achieves the lowest MIA on TOFU ($17.8$ vs.\ our $18.0$). 
These pockets of strength align with capacity and regularization differences (full-FT’s higher capacity benefits single-aspect unlearning; LoRA-based variants can further suppress leakage), yet \ourmethod remains Pareto-favorable on the aggregate.

\textbf{Obs 2.} \textbf{\ourmethod preserves general utility while delivering strong, robust unlearning.} 
Across datasets, \ourmethod’s GUR is uniformly the best, indicating minimal collateral damage to non-target capabilities (e.g., EDU-RELAT $95.1\%$ and KnowUnDo $95.6\%$). 
At the same time, APR, which is our adversarial robustness proxy, is best or near-best in three datasets (e.g., EDU-RELAT $82.3\%$, RWKU $79.4\%$, TOFU $80.8\%$), with only a marginal gap on KnowUnDo (Yao-Neg $84.2\%$ vs.\ ours $83.9\%$). 
This pattern supports the hypothesis that constraining edits to low-rank adapters focuses updates on the intended behaviors, mitigating over-unlearning and improving robustness to prompt variants.

\textbf{Obs 3.} \textbf{Capacity and regularization effects explain remaining gaps.} 
Where full-parameter updates excel (e.g., EDU-RELAT USR $91.6\%$, KnowUnDo APR $84.2\%$ for Yao-Neg), gains are concentrated on a single axis, often accompanied by weaker generality or privacy relative to \ourmethod (e.g., GUR and MIA). 
Conversely, LoKU’s lowest TOFU MIA ($17.8\%$) highlights the privacy advantage of low-rank updates, yet \ourmethod still balances leakage with superior utility and robustness (higher GUR/APR in the same table). 
Overall, \ourmethod consistently shifts the utility-unlearning-privacy frontier outward.

\textbf{Obs 4.} \textbf{LoRA-aware design matters beyond simply “using LoRA.”} 
To isolate this factor, \Cref{tab:yao_lora_vs_ours} compares \ourmethod with Yao-Neg (LoRA) under the same negative-only objective. 
\ourmethod outperforms Yao-Neg (LoRA) on all datasets and metrics (e.g., EDU-RELAT GUR $95.1\%$ vs.\ $91.1\%$, RWKU USR $88.5\%$ vs.\ $83.7\%$, KnowUnDo MIA $17.2\%$ vs.\ $24.2\%$, TOFU APR $80.8\%$ vs.\ $75.5\%$). 
We attribute this consistent margin to our LoRA-specific choices (module selection, rank scanning, early stopping) and multi-view negative construction, which together localize edits and stabilize general utility while maintaining robust unlearning and low leakage.

\begin{figure}[t]
    \centering
    \subfloat[{EDU-RELAT}]{\includegraphics[height=1.27in]{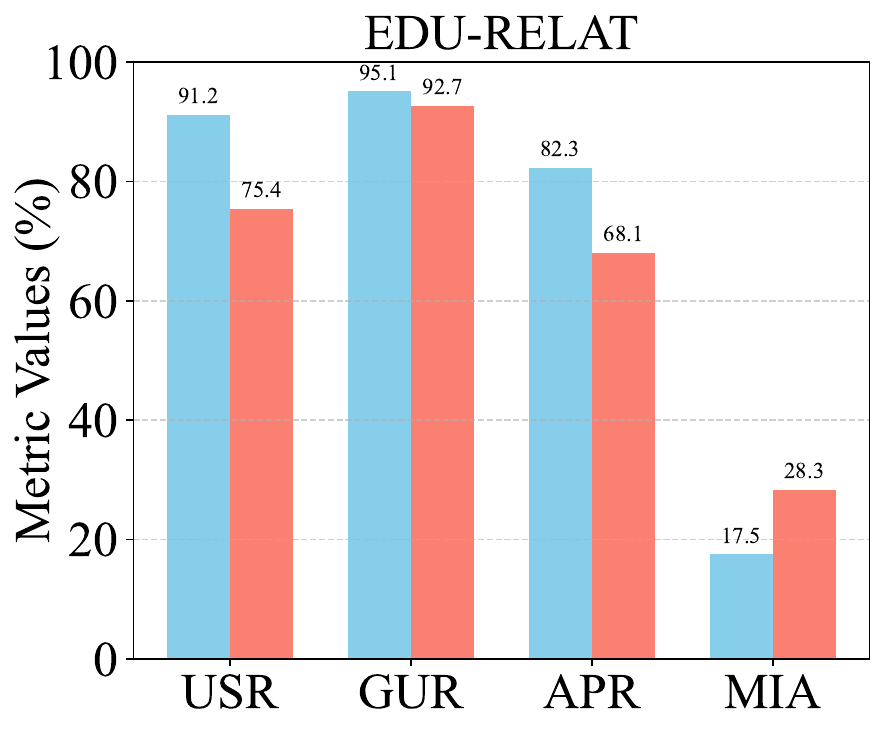}
    }\hspace{-0.5em}
    \subfloat[{RWKU}]{\includegraphics[height=1.27in]{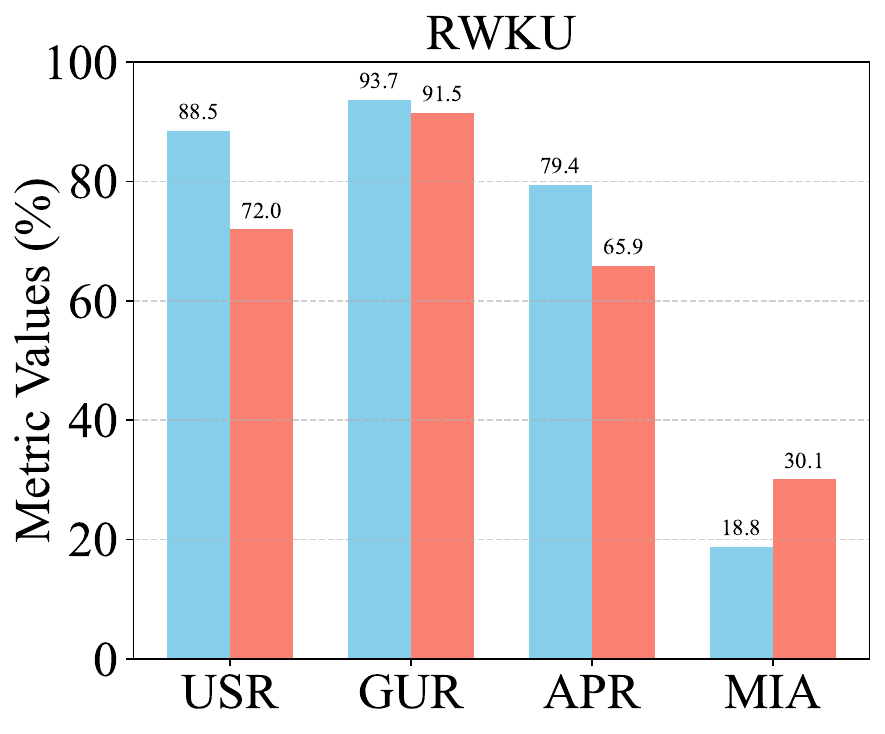}
    }\hspace{-0.5em}
    \vspace{1em}
    \subfloat[{KnowUnDo}]{\includegraphics[height=1.27in]{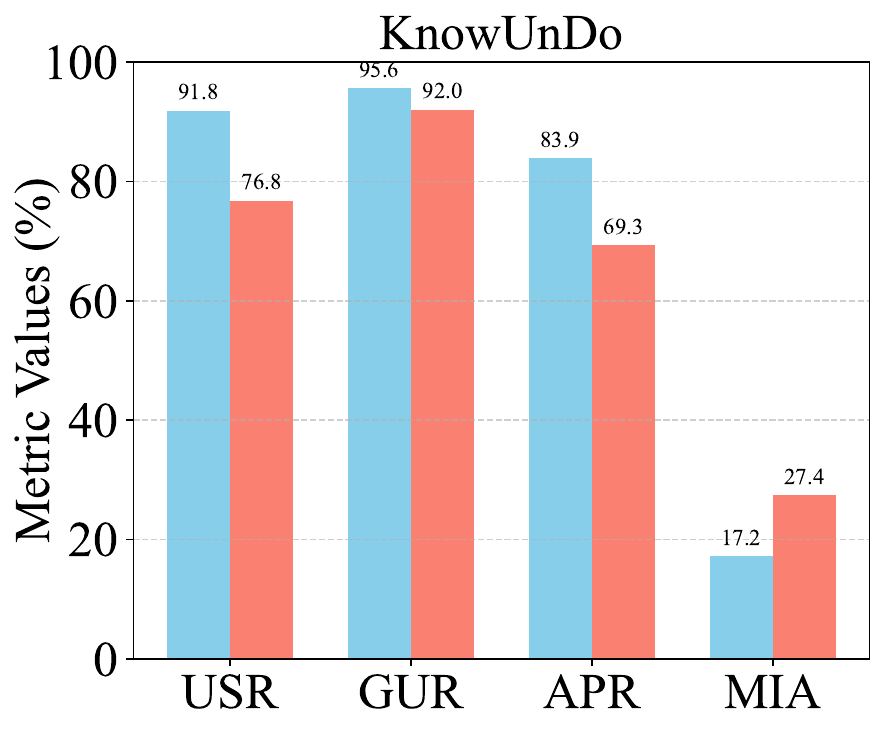}\label{fair:test1}
    }\hspace{-0.5em}
    \subfloat[{TOFU}]{\includegraphics[height=1.27in]{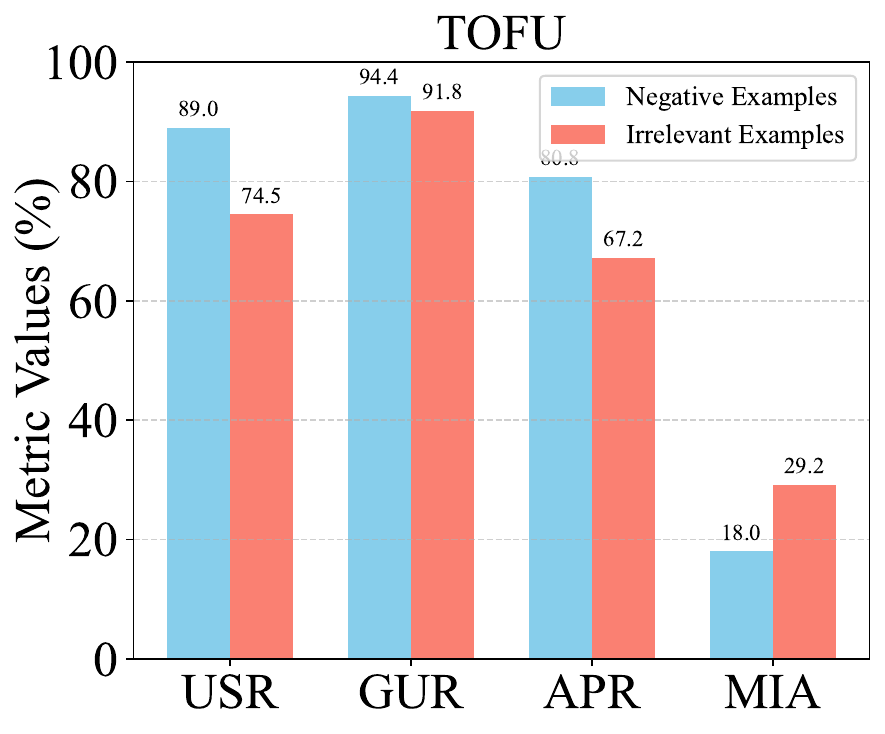}
    }
    \vspace{-1.2em}
    \caption{\textbf{Ablation study: negative vs. irrelevant examples.}  
Fine-tuning with negative examples leads to higher unlearning effectiveness and robustness ($\uparrow$USR, $\uparrow$APR, $\downarrow$MIA), while maintaining high utility ($\uparrow$GUR), consistently outperforming irrelevant examples across all datasets.}
    \label{fig:distribution}
    \vspace{-1.5em}
\end{figure}

\subsection{Ablation Study: Negative vs. Irrelevant Examples}
To further understand what drives effective unlearning, we conduct an ablation study comparing fine-tuning with explicitly constructed negative examples (used in \textbf{{\ourmethod}}) against a baseline using randomly selected irrelevant examples. This comparison aims to answer: \textit{How can unlearning be achieved efficiently and robustly without full retraining?}

Across all four datasets, we observe consistent and significant performance gains when using negative examples. Specifically, negative-example fine-tuning achieves a higher unlearning success rate ($\uparrow$USR) and adversarial probe rejection ($\uparrow$APR), while lowering membership inference attack success ($\downarrow$MIA), all with minimal compromise to general utility ($\uparrow$GUR). \textbf{(i)} Negative examples provide stronger contrastive supervision, guiding parameter updates more effectively toward forgetting specific information. \textbf{(ii)} Random irrelevant examples lack semantic opposition to the target knowledge, thus failing to generate useful gradients for unlearning. \textbf{(iii)} The effectiveness of counterfactual-style supervision generalizes well across domains and tasks, validating the robustness of our strategy. These findings support the design choice in \textbf{{\ourmethod}} to incorporate targeted negative supervision and highlight the importance of principled data construction for reliable and efficient unlearning.

\vspace{-0.3em}
\subsection{The Effect of Low-Rank $r$}
\begin{figure}[t]
    \centering
    \subfloat[{EDU-RELAT}]{\includegraphics[height=1.27in]{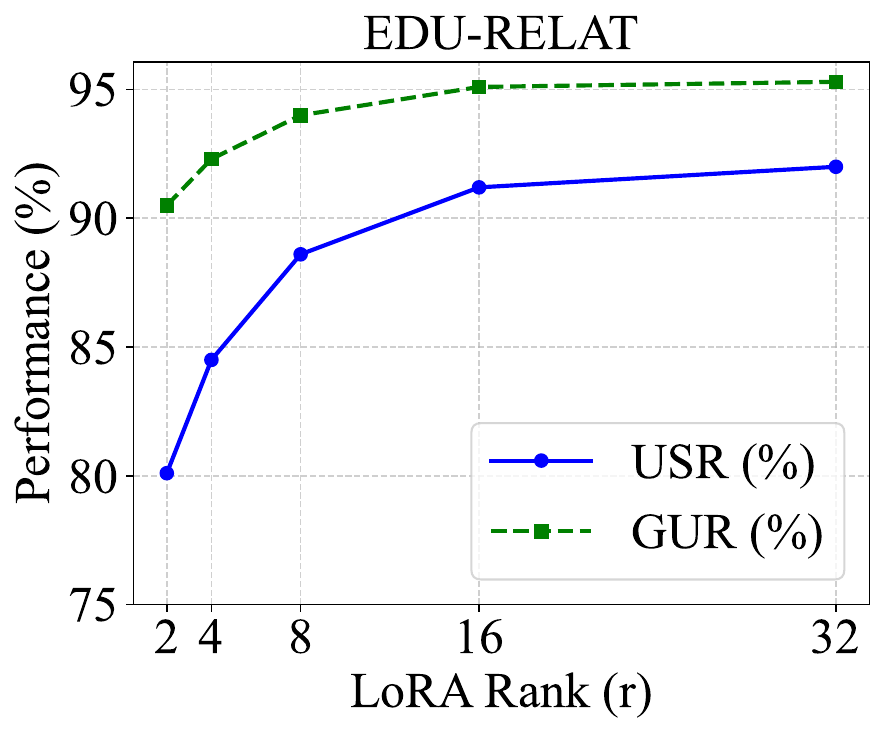}
    }\hspace{-0.5em}
    \subfloat[{RWKU})]{\includegraphics[height=1.27in]{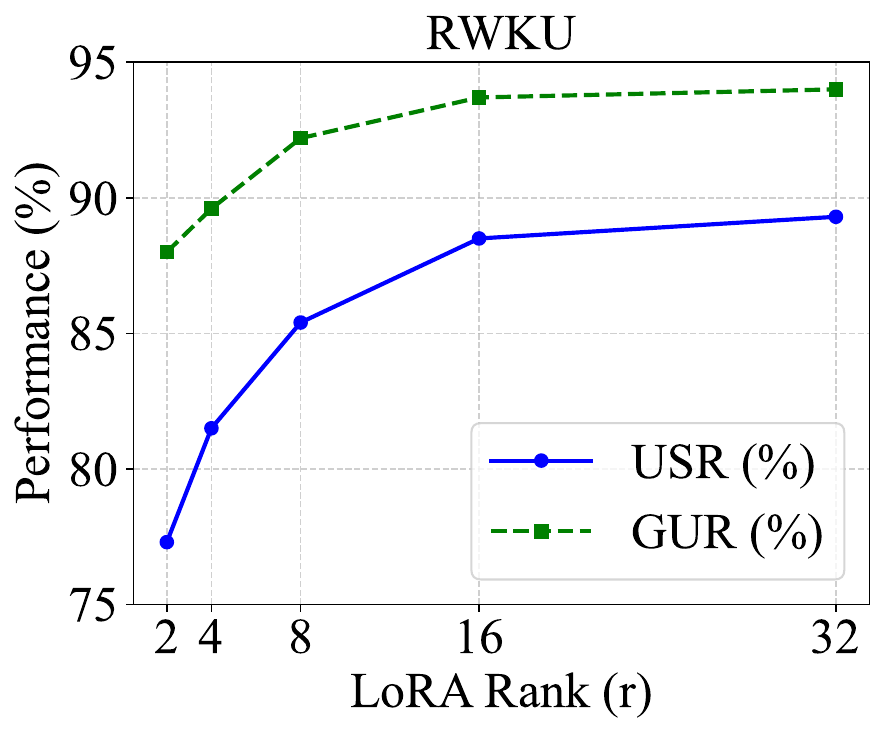}
    }\hspace{-0.5em}
    \vspace{1em}
    \subfloat[{KnowUnDo}]{\includegraphics[height=1.27in]{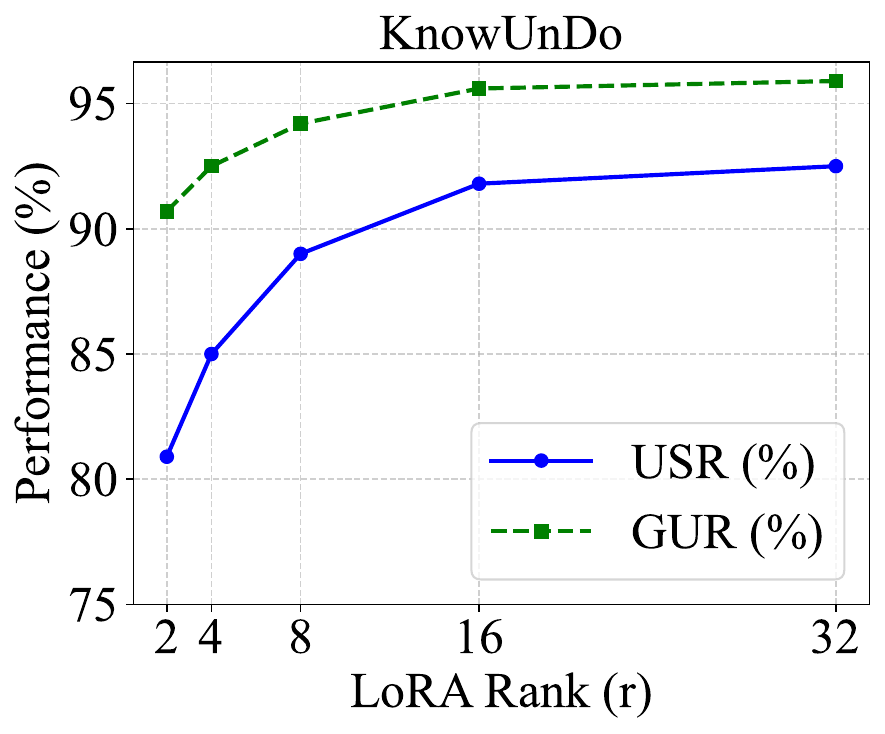}\label{fair:test1}
    }\hspace{-0.5em}
    \subfloat[{TOFU}]{\includegraphics[height=1.27in]{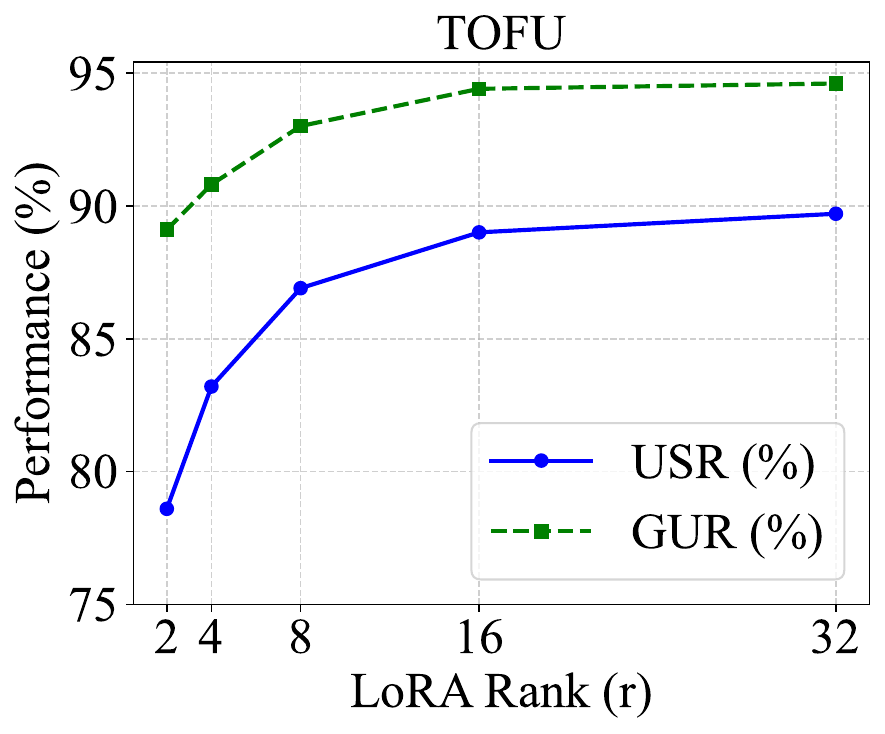}
    }
        \vspace{-1.2em}
    \caption{\textbf{The effect of low-rank $r$}. Performance improves with larger $r$ up to 16, after which gains saturate. Moderate ranks ({\eg}, $r=8$ or $r=16$) offer the best trade-off.}
    \label{fig:lora_rank}
\end{figure}

\vspace{-0.3em}
To evaluate how the capacity of LoRA adapters influences unlearning effectiveness, we conduct an ablation study by varying the LoRA rank $r \in {2, 4, 8, 16, 32}$ across all four datasets. As shown in~\Cref{fig:lora_rank}, increasing the rank generally improves both unlearning success rate (USR) and general utility retention (GUR), with the most notable gains occurring between ranks 2 and 16. Beyond $r=16$, the performance tends to plateau, suggesting diminishing returns with higher parameter capacity. This trend is consistent across datasets, highlighting that moderate-rank LoRA ({\eg}, $r=8$ or $r=16$) provides an optimal trade-off between effectiveness and efficiency. These findings reinforce the practicality of {\ourmethod} in resource-constrained settings, where low-rank adaptation enables effective unlearning with minimal computational overhead.

As in~\Cref{fig:lora_rank}, increasing the LoRA rank generally improves both unlearning success rate ($\uparrow$USR) and general utility retention ($\uparrow$GUR), especially in the range from $r=2$ to $r=16$. This suggests that higher-rank adapters provide greater expressive capacity to capture the negative gradients necessary for effective unlearning. However, performance gains diminish beyond $r=16$, and further increases (e.g., $r=32$) incur additional computation with only marginal improvements. In fact, extremely high-rank adaptation may lead to overfitting or instability in certain cases. (i) A moderate LoRA rank (e.g., $r=8$ or $r=16$) offers an optimal trade-off between unlearning effectiveness and training efficiency. In contrast, (ii) very low ranks ($r=2$ or $r=4$) underperform, resulting in degraded USR and slightly reduced GUR, likely due to insufficient parameter capacity to accommodate meaningful updates. These observations remain consistent across all datasets, despite differences in domain and task structure. (iii) The rank-performance relationship is stable and generalizable, reinforcing the robustness of {\ourmethod}’s design. (iv) {\ourmethod} maintains competitive performance even under low-rank settings, which highlights its practicality for computationally constrained scenarios, making it a viable unlearning solution for large-scale LLMs with limited resources.

\begin{table*}[t]
\centering
\small
\setlength{\tabcolsep}{6pt}
\renewcommand{\arraystretch}{1.2}

\begin{tabular}{p{0.15\textwidth} p{0.40\textwidth} p{0.35\textwidth}}
\Xhline{1.2pt}
\rowcolor{tablehead!20}
\textbf{Dataset} & \textbf{Original Fact} & \textbf{Alternative Negative Example} \\
\hline\hline
\rowcolor{gray!10}\textbf{EDU-RELAT} & John's brother is Mike. & John's brother is Kevin. \\
\textbf{RWKU} & The capital of France is Paris. & The capital of France is Lyon. \\
\rowcolor{gray!10}\textbf{KnowUnDo} & Alice works at Microsoft. & Alice works at Google. \\
\multirow{2}{*}{\textbf{TOFU}} & Author Alex Smith primarily writes science fiction novels. & Author Alex Smith primarily writes romance novels. \\
\Xhline{1.2pt}
\end{tabular}

\caption{Negative examples generated by replacing true facts with alternative (erroneous) information for each dataset.}
\label{tab:negative_samples_alternative}
\end{table*}
\subsection{Quality of the Example}
To evaluate the impact of negative example quality on unlearning performance, we construct three variants of training data representing different levels of semantic clarity. 
\noindent\textbf{High-quality examples} are explicit contradictions of the target knowledge, such as \textit{``The capital of France is Lyon''}, which directly oppose the fact to be unlearned. 

\noindent\textbf{Medium-quality examples} introduce subtle ambiguity or hedging without fully committing to a contradictory statement, for example, \textit{``Paris may not always be considered France’s capital''}. These examples may confuse the model rather than guiding it to forget. 

\noindent\textbf{Low-quality examples} are loosely related or entirely irrelevant statements, such as \textit{``France has many important cities''}, which lack any clear corrective signal. By fine-tuning {\ourmethod} on each variant independently, we assess how the clarity of the negative supervision affects targeted unlearning, general knowledge retention, and robustness to adversarial prompts.

The results in~\Cref{tab:quality_negative_samples} demonstrate that the quality of negative examples plays a pivotal role in the performance of {\ourmethod}. Specifically, we observe a strong correlation between the clarity and specificity of negative examples and the model’s unlearning effectiveness and robustness.

\begin{fullitemize}
\item USR: High-quality negative examples lead to a significantly higher USR (88.5\%) compared to medium (78.0\%) and low-quality (65.3\%) examples. This indicates that explicitly contradicting the undesired information is essential for effectively suppressing the model’s prior knowledge. When the negative example is vague or only weakly related, the model struggles to identify which behavior to unlearn.

\item APR: A similar trend is observed in the model's robustness to paraphrased prompts. With high-quality examples, APR reaches 79.4\%, but drops significantly with medium (67.2\%) and low-quality (53.9\%) examples. This suggests that only clear and direct contradictions can generalize well to variations in user input.

\item GUR: All example types maintain relatively high GUR, though high-quality examples preserve it best (93.7\%), slightly outperforming medium (91.0\%) and low-quality (89.4\%) examples. Notably, poorly crafted examples can interfere with unrelated knowledge, leading to a minor degradation in general performance due to noisier gradient updates.

\item MIA Accuracy: Lower MIA accuracy with high-quality examples (18.8\%) reflects more successful removal of memorized facts. In contrast, higher MIA accuracy for low-quality examples (32.0\%) implies that vague or unrelated examples fail to overwrite the memorized information, leaving the model vulnerable to inference attacks.
\end{fullitemize}
\begin{table}[t]
\centering
\small
\renewcommand{\arraystretch}{1.0}
\begin{tabular}{lcccc}
\Xhline{1.2pt}
\rowcolor{tablehead!20}
\textbf{Neg. Examples Quality} & \textbf{USR (\%)} & \textbf{APR (\%)} & \textbf{GUR (\%)} & \textbf{MIA (\%)} \\
\hline\hline
\rowcolor{gray!10}High Quality   & $88.5$ & $79.4$ & $93.7$ & $18.8$ \\
Medium Quality & $78.0$ & $67.2$ & $91.0$ & $26.5$ \\
\rowcolor{gray!10}Low Quality    & $65.3$ & $53.9$ & $89.4$ & $32.0$ \\
\Xhline{1.2pt}
\end{tabular}
\caption{\textbf{Impact of negative example quality} on the RWKU dataset. ``Neg.'' is abbreviated for Negative. Higher-quality negative examples yield stronger unlearning effectiveness and robustness while preserving utility, demonstrating the importance of clarity and specificity in negative supervision.}
\label{tab:quality_negative_samples}
\vspace{-1em}
\end{table}
\section{Conclusion}\label{sec:impact}
We present \textbf{\ourmethod}~\protect\includegraphics[scale=0.12,valign=c]{figure/LUNE_logo.pdf}, a LoRA-based, negative-only unlearning framework that edits \emph{only} lightweight adapters to remove targeted knowledge efficiently. 
By localizing updates, \ourmethod suppresses undesired behaviors while preserving general ability, mitigating catastrophic drift. 
Across four benchmarks, it achieves strong unlearning (USR/APR) with superior utility retention (GUR) and lower leakage (MIA). 
Future work includes improving cross-task generalization and extending to concept- and multi-instance unlearning.

\section*{Acknowledgements}
This work was supported in part by the DARPA Young Faculty Award, the National Science Foundation (NSF) under Grants \#2127780, \#2319198, \#2321840, \#2312517, and \#2235472, \#2431561, the Semiconductor Research Corporation (SRC), the Office of Naval Research through the Young Investigator Program Award, and Grants \#N00014-21-1-2225 and \#N00014-22-1-2067, Army Research Office Grant \#W911NF2410360. Additionally, support was provided by the Air Force Office of Scientific Research under Award \#FA9550-22-1-0253, along with generous gifts from Xilinx and Cisco.
\bibliographystyle{plain}
\bibliography{ref}

\newpage
\appendix
\section{Detailed Introduction to Baselines}\label{app:baseline-details}
\begin{fullitemize}
\item \textbf{Gradient Ascent (GA)}~\citep{jang2023knowledge} 
This baseline utilizes gradient ascent to maximize the model's loss on the target information, explicitly discouraging the model from generating undesired content. While simple, GA is computationally expensive and may negatively impact unrelated knowledge due to aggressive updates.

\item  \textbf{NPO.}~\citep{zhang2024negative}
Frames unlearning as \emph{preference optimization} against negative responses: the model is trained to prefer safe/non-target outputs over negative ones.
Compared with naive negative-only FT, NPO usually offers a more stable trade-off between unlearning and utility, given well-constructed preference pairs.

\item \textbf{Task Vector (TV).}~\citep{ilharco2022editing} 
Task Vector identifies a specific direction in the parameter space of the pretrained model corresponding to the target knowledge. Unlearning is achieved by subtracting this direction from the model parameters. While efficient, it can unintentionally affect semantically related knowledge.

\item \textbf{MemFlex.}~\citep{tian2024forget}
MemFlex employs gradient-based optimization to precisely remove sensitive parameters associated with undesired information. This method maintains general knowledge but demands access to gradient computations across large parameter subsets, increasing computational complexity.

\item  \textbf{Yao-Neg.}~\citep{yao2024large}
Unlearning is performed by \emph{fine-tuning only on negative examples}, updating all model parameters to suppress targeted knowledge/behaviors. In the main experiment (\Cref{tab:evaluation_results}), we report \emph{Yao–Neg} in its \emph{full fine-tuning} configuration, which the original paper presents as a primary/standard instantiation alongside an optional LoRA variant; for completeness, results for the \emph{LoRA} version appear in the appendix (\Cref{tab:yao_lora_vs_ours}), computed under the same negative-only setup.

\item  \textbf{LoKU.}~\citep{cha2024towards}
A \emph{LoRA-based} unlearning approach that freezes the backbone and trains low-rank adapters, often with stabilizing regularizers. It is parameter-efficient and localizes edits, typically retaining utility better and reducing leakage compared to full-parameter updates.

\end{fullitemize}
\section{Detailed Setups}\label{app:setup}
\subsection{Training/Fine-tuning Setups}
Unless otherwise noted, we train \ourmethod with \emph{negative-only} supervision using the same data splits and evaluation protocol as the main text, and keep all backbone weights frozen while updating only LoRA adapters. We apply LoRA to attention projections ($W_q,W_k,W_v,W_o$) and the FFN up/down projections, initialize adapters to zero, and use a default rank $r{=}16$ (chosen from an ablation over $r\in\{2,4,8,16,32\}$ where gains saturate near $16$); LoRA dropout is $0.05$ and scaling $\alpha{=}r$. Optimization uses AdamW with learning rate $2\!\times\!10^{-4}$, linear warmup over the first 5\% of steps, cosine decay thereafter, weight decay $0.01$, gradient clipping at $1.0$, mixed-precision (bf16), and gradient accumulation to match an effective batch size of $256$ tokens/step. Inputs are tokenized with the backbone tokenizer; we cap sequence length at $1{,}024$ for training and evaluation, pad on the right, and mask loss to target spans only. Early stopping on a held-out negative-dev set monitors USR$\uparrow$/APR$\uparrow$ at fixed GUR$\uparrow$ tolerance ($\le 0.5$pp drop). For fairness across methods, we keep the total \#steps per dataset aligned to the epoch budgets reported in the paper (\Cref{tab:datasets_models,tab:epochs_per_dataset}) and repeat each run with three random seeds, reporting mean~$\pm$~standard error of the mean for all metrics.

\subsection{LLM Backbones}\label{app:llm_backbone}
\paragraph{\llmname{Mistral 7B}.} It is a 7b-parameter LLM by Mistral AI~\citep{jiang2023mistral7b}, effectively handles text generation and diverse NLP tasks, whose benchmark covers areas like commonsense reasoning, world knowledge, math, and reading comprehension, showcasing its broad applicability. It utilizes a sliding window attention mechanism~\citep{child2019generating,beltagy2020longformer}, supports English and coding languages, and operates with an 8k context length. 

\paragraph{\llmname{LLaMA-2 7B}.} \llmname{LLaMA-2} (Large Language Model Meta AI) is an open-source family of autoregressive transformer-based language models released by Meta AI~\citep{touvron2023llama}. The 7B variant offers a strong balance between performance and computational efficiency, making it a practical choice for fine-tuning and unlearning experiments on consumer-level hardware. Pretrained on a diverse mix of publicly available data, \llmname{LLaMA-2 7B} exhibits competitive language understanding and generation capabilities compared to larger proprietary models, while remaining accessible for research and reproducibility. 

\begin{table*}[t]
\centering
\small
\setlength{\tabcolsep}{6pt}
\renewcommand{\arraystretch}{1.0}

\begin{tabular}{lccc}
\Xhline{1.2pt}
\rowcolor{tablehead!20}
\textbf{Dataset} & \textbf{\# Samples} & \textbf{Epochs} & \textbf{Task} \\
\hline\hline
\rowcolor{gray!10}\textbf{EDU-RELAT}  & $10{,}000$ & $30$ & Synthetic relational data; quick convergence \\
\textbf{RWKU}       & $13{,}000$ & $40$ & Larger real-world factual dataset \\
\rowcolor{gray!10}\textbf{KnowUnDo}   & $8{,}000$  & $35$ & Entity privacy-sensitive task \\
\textbf{TOFU}       & $4{,}000$  & $50$ & Smaller profile data; longer tuning needed \\
\Xhline{1.2pt}
\end{tabular}
\caption{Number of training epochs used for {\ourmethod} on each dataset. The values are selected based on dataset size and convergence behavior.}
\label{tab:epochs_per_dataset}
\end{table*}

\subsection{Negative Examples Generation}\label{app:gen}
To effectively guide the unlearning process, we explicitly generate negative examples designed to counteract previously learned, undesired knowledge. ~\Cref{tab:negative_samples_alternative} provides illustrative examples of alternative negative examples tailored specifically for each dataset used in our evaluation. The generation of negative examples follows these key steps:
\begin{fullitemize}
    \item \textbf{Identify Target Knowledge}:  
    Clearly define the specific facts, associations, or behaviors that the model should unlearn.
    
    \item \textbf{Construct Contradictory or Alternative Statements}:  
    Create statements that explicitly contradict or replace the targeted information. These statements can be either directly contradictory ({\eg}, \textit{``The capital of France is not Paris''}) or alternative erroneous facts ({\eg}, \textit{``The capital of France is Lyon''}).

    \item \textbf{Paraphrase and Diversify Examples}:  
    Generate multiple paraphrased variations to enhance robustness and generalization of the unlearning effect across different prompt forms and phrasings.

    \item \textbf{Validate and Curate}:  
    Verify that the negative examples clearly and effectively negate or overwrite the undesired knowledge without introducing unintended biases or misinformation beyond the targeted scope.
\end{fullitemize}

\section{More Experiments}
\begin{table}[t]
\centering
\small
\setlength{\tabcolsep}{6pt}
\renewcommand{\arraystretch}{1.0}

\begin{tabular}{ll|cccc}
\Xhline{1.2pt}
\rowcolor{tablehead!20}
\textbf{Dataset} & \textbf{Method} & \textbf{USR (\%)} & \textbf{GUR (\%)} & \textbf{APR (\%)} & \textbf{MIA (\%)} \\
\hline\hline
\rowcolor{gray!10}& Full FT            & $88.7$ & $90.2$ & $79.5$ & $25.4$ \\
\multirow{-2}{*}{\textbf{EDU-RELAT}} 
                                  & \textbf{{\ourmethod}} & $\mathbf{91.2}$ & $\mathbf{95.1}$ & $\mathbf{82.3}$ & $\mathbf{17.5}$ \\
\hline
\rowcolor{gray!10}   & Full FT            & $85.1$ & $89.4$ & $76.0$ & $27.8$ \\
   \multirow{-2}{*}{\textbf{RWKU}}   
                                    & \textbf{{\ourmethod}} & $\mathbf{88.5}$ & $\mathbf{93.7}$ & $\mathbf{79.4}$ & $\mathbf{18.8}$ \\
\hline
\rowcolor{gray!10}& Full FT            & $89.0$ & $91.5$ & $80.2$ & $24.1$ \\
\multirow{-2}{*}{\textbf{KnowUnDo}}
                                    & \textbf{{\ourmethod}} & $\mathbf{91.8}$ & $\mathbf{95.6}$ & $\mathbf{83.9}$ & $\mathbf{17.2}$ \\
\hline
\rowcolor{gray!10}      & Full FT            & $86.4$ & $89.9$ & $77.3$ & $26.5$ \\
      \multirow{-2}{*}{\textbf{TOFU}}
                                    & \textbf{{\ourmethod}} & $\mathbf{89.0}$ & $\mathbf{94.4}$ & $\mathbf{80.8}$ & $\mathbf{18.0}$ \\
                                
\Xhline{1.2pt}
\end{tabular}

\caption{Ablation study comparing LoRA-based fine-tuning ({\ourmethod}) and full fine-tuning (FT) on all parameters across four datasets. Metrics shown include Unlearning Success Rate (USR), General Utility Retention (GUR), Adversarial Probe Rejection (APR), and Membership Inference Attack accuracy (MIA). The best results are highlighted in bold.}
\label{tab:lora_vs_full}
\end{table}

\subsection{Comparison of LoRA vs. Full Fine-Tuning}
To assess the efficiency and effectiveness of our proposed method, we compare {\ourmethod}, which fine-tunes only a small set of LoRA adapters, with traditional full fine-tuning that updates all model parameters. As shown in~\Cref{tab:lora_vs_full}, {\ourmethod} consistently achieves comparable or superior performance across all evaluation metrics. Specifically, {\ourmethod} outperforms full fine-tuning in unlearning success rate (USR) and adversarial robustness (APR), while also maintaining higher general utility (GUR) and achieving lower MIA accuracy, indicating improved privacy. These results highlight that LoRA-based adaptation not only reduces computational cost but also enables more targeted and reliable unlearning, making it a more practical and scalable approach for real-world applications.

\subsection{Additional Comparison}

We additionally instantiate Yao et al.'s negative-only unlearning with \emph{LoRA} adapters and report its results separately in~\Cref{tab:yao_lora_vs_ours}. 
Unless noted, we reuse the same datasets, negative-example construction, and training budgets as in~\Cref{tab:evaluation_results}. 
The only change is the optimization regime: instead of full-parameter fine-tuning used by \emph{Yao-Neg (Full-FT)} in the main table, we enable LoRA with the standard target modules (attention/FFN) and the same rank/search protocol as our method, while keeping all other hyperparameters identical. 
This isolates the effect of using LoRA under the \emph{same} negative-only objective and makes the comparison to our \emph{LoRA-only} design fair and transparent.

\begin{table}[t]
\centering
\small
\renewcommand{\arraystretch}{1.0}

\begin{tabular}{ll|cc}
\Xhline{1.2pt}
\rowcolor{tablehead!20}
\textbf{Dataset} & \textbf{Metric} & \textbf{Yao\text{-}Neg (LoRA)} & \textbf{{\ourmethod} (Ours)} \\
\hline\hline
\rowcolor{gray!10}& USR (\%) & $89.3\pm0.4$ & $\mathbf{91.2\pm0.3}$ \\
& GUR (\%) & $91.1\pm0.3$ & $\mathbf{95.1\pm0.2}$ \\
\rowcolor{gray!10}& APR (\%) & $78.0\pm0.4$ & $\mathbf{82.3\pm0.3}$ \\
\multirow{-4}{*}{\textbf{EDU-RELAT}}
& MIA (\%) & $23.6\pm0.3$ & $\mathbf{17.5\pm0.2}$ \\
\hline
\rowcolor{gray!10}& USR (\%) & $83.7\pm0.4$ & $\mathbf{88.5\pm0.3}$ \\
& GUR (\%) & $88.6\pm0.3$ & $\mathbf{93.7\pm0.2}$ \\
\rowcolor{gray!10}& APR (\%) & $74.8\pm0.5$ & $\mathbf{79.4\pm0.3}$ \\
\multirow{-4}{*}{\textbf{RWKU}}
& MIA (\%) & $26.0\pm0.3$ & $\mathbf{18.8\pm0.2}$ \\
\hline
\rowcolor{gray!10}& USR (\%) & $87.4\pm0.4$ & $\mathbf{91.8\pm0.3}$ \\
& GUR (\%) & $92.0\pm0.3$ & $\mathbf{95.6\pm0.2}$ \\
\rowcolor{gray!10}& APR (\%) & $80.1\pm0.4$ & $\mathbf{83.9\pm0.3}$ \\
\multirow{-4}{*}{\textbf{KnowUnDo}}
& MIA (\%) & $24.2\pm0.3$ & $\mathbf{17.2\pm0.2}$ \\
\hline
\rowcolor{gray!10}& USR (\%) & $84.2\pm0.4$ & $\mathbf{89.0\pm0.3}$ \\
& GUR (\%) & $90.3\pm0.3$ & $\mathbf{94.4\pm0.2}$ \\
\rowcolor{gray!10}& APR (\%) & $75.5\pm0.4$ & $\mathbf{80.8\pm0.3}$ \\
\multirow{-4}{*}{\textbf{TOFU}}
& MIA (\%) & $24.9\pm0.3$ & $\mathbf{18.0\pm0.2}$ \\
\Xhline{1.2pt}
\end{tabular}

\vspace{1em}
\caption{Comparison between Yao\text{-}Neg (LoRA) and our method across datasets. Best results in \textbf{bold}. (For MIA, lower is better.)}
\label{tab:yao_lora_vs_ours}
\vspace{-1em}
\end{table}

Across all datasets and metrics in~\Cref{tab:yao_lora_vs_ours}, our method consistently outperforms \emph{Yao-Neg (LoRA)}. 
We attribute this to three factors. 
(i) \textbf{LoRA-aware design}: our training is tailored to low-rank adapters (module selection, rank scanning, and early stopping), which localizes edits and stabilizes utility, whereas \emph{Yao-Neg (LoRA)} directly ports the negative-only objective without LoRA-specific regularization. 
(ii) \textbf{Robust negatives}: our multi-view negative construction (paraphrase/counterfactual/retrieval variants) improves robustness, yielding higher USR/APR under the same budget. 
(iii) \textbf{Drift control}: by freezing the backbone and constraining updates to low-rank adapters, our method reduces unintended drift (reflected by higher GUR and lower MIA).
Notably, the gains hold on both medium-scale (\textsc{EDU-RELAT}, \textsc{RWKU}) and large-scale datasets (\textsc{KnowUnDo}, \textsc{TOFU}), indicating that our LoRA-specific design scales while preserving strong unlearning-utility trade-offs.

\section{Additional Discussion}
To provide a balanced view, we complement the earlier discussion of limitations with a brief account of our method’s merits in this Appendix.

The proposed \textbf{{\ourmethod}} method introduces a lightweight and targeted approach to unlearning in LLMs, offering several key advantages. First, it is highly \textit{parameter-efficient}, as it fine-tunes only the low-rank LoRA matrices \( A \) and \( B \), drastically reducing the number of trainable parameters compared to full model fine-tuning. This not only lowers computational and memory demands but also makes {\ourmethod} practical in resource-constrained settings. Second, {\ourmethod} ensures \textit{preservation of original knowledge} by freezing the pre-trained model weights \( W_0 \), thereby maintaining the model’s general capabilities and minimizing the risk of catastrophic unlearning. Third, {\ourmethod} performs \textit{targeted unlearning} by fine-tuning exclusively on negative examples, allowing the model to forget specific information without requiring access to the full training set, a valuable feature when data availability is limited or privacy-sensitive. Altogether, {\ourmethod} offers an effective, scalable, and focused solution for unlearning in LLMs by combining LoRA’s parameter efficiency with task-specific negative supervision.

\end{document}